\pdfoutput=1

\documentclass[11pt]{article}

\usepackage[final]{acl}

\usepackage{times}
\usepackage{latexsym}

\usepackage[T1]{fontenc}

\usepackage[utf8]{inputenc}

\usepackage{microtype}

\usepackage{inconsolata}

\usepackage{graphicx}

\usepackage{bibentry}
\usepackage{hyperref}
\usepackage{tabularx}
\usepackage{algpseudocode}
\usepackage{amsmath}
\usepackage{cite} 
\usepackage{array}
\usepackage{placeins}
\usepackage{float}
\usepackage{booktabs} 
\usepackage{helvet}  
\usepackage{courier}  
\usepackage{graphicx} 
\usepackage{natbib}  
\usepackage{caption} 
\usepackage{algorithm}
\usepackage[utf8]{inputenc}
\usepackage{CJKutf8}

\title{LawLuo: A Multi-Agent Collaborative Framework for Multi-Round Chinese Legal Consultation}
\author{
   Jingyun Sun$^1$, Chengxiao Dai$^2$, Zhongze Luo$^1$, Yangbo Chang$^3$, Yang Li$^{1, *}$ \\
   College of Computer and Control Engineering, Northeast Forestry University\\
Faculty of Information and Communication Technology, Universiti Tunku Abdul Rahman\\ 
College of Architecture and Arts Design, Shijiazhuang Tiedao University
}

\begin{document}
\maketitle
\begin{abstract}
Legal Large Language Models (LLMs) have shown promise in providing legal consultations to non-experts. However, most existing Chinese legal consultation models are based on single-agent systems, which differ from real-world legal consultations, where multiple professionals collaborate to offer more tailored responses. \textbf{To better simulate real consultations, we propose LawLuo, a multi-agent framework for multi-turn Chinese legal consultations}. LawLuo includes four agents: the receptionist agent, which assesses user intent and selects a lawyer agent; the lawyer agent, which interacts with the user; the secretary agent, which organizes conversation records and generates consultation reports; and the boss agent, which evaluates the performance of the lawyer and secretary agents to ensure optimal results. These agents’ interactions mimic the operations of real law firms. To train them to follow different legal instructions, we developed distinct fine-tuning datasets. We also introduce a case graph-based RAG to help the lawyer agent address vague user inputs. Experimental results show that LawLuo outperforms baselines in generating more personalized and professional responses, handling ambiguous queries, and following legal instructions in multi-turn conversations. Our full code and constructed datasets will be open-sourced upon paper acceptance.
\end{abstract}

%

\section{Introduction}
\label{section:intro}

Since the release of ChatGPT, the development of Chinese Large Language Models (LLMs) has accelerated, resulting in influential models like ChatGLM \citep{glm}, LLaMa \citep{llama}, and BaiChuan \citep{baichuan}. These models excel in fluent Chinese dialogue and understanding complex user intentions. Additionally, domain-specific LLMs, such as Medical \citep{zhongjing, huatuo}, Legal \citep{lawgpt, lawyerllama}, and Financial LLMs \citep{xuanyuan}, have emerged, demonstrating strong capabilities in their respective fields.

\begin{figure*}
    \centering
    \includegraphics[width=0.9\linewidth]{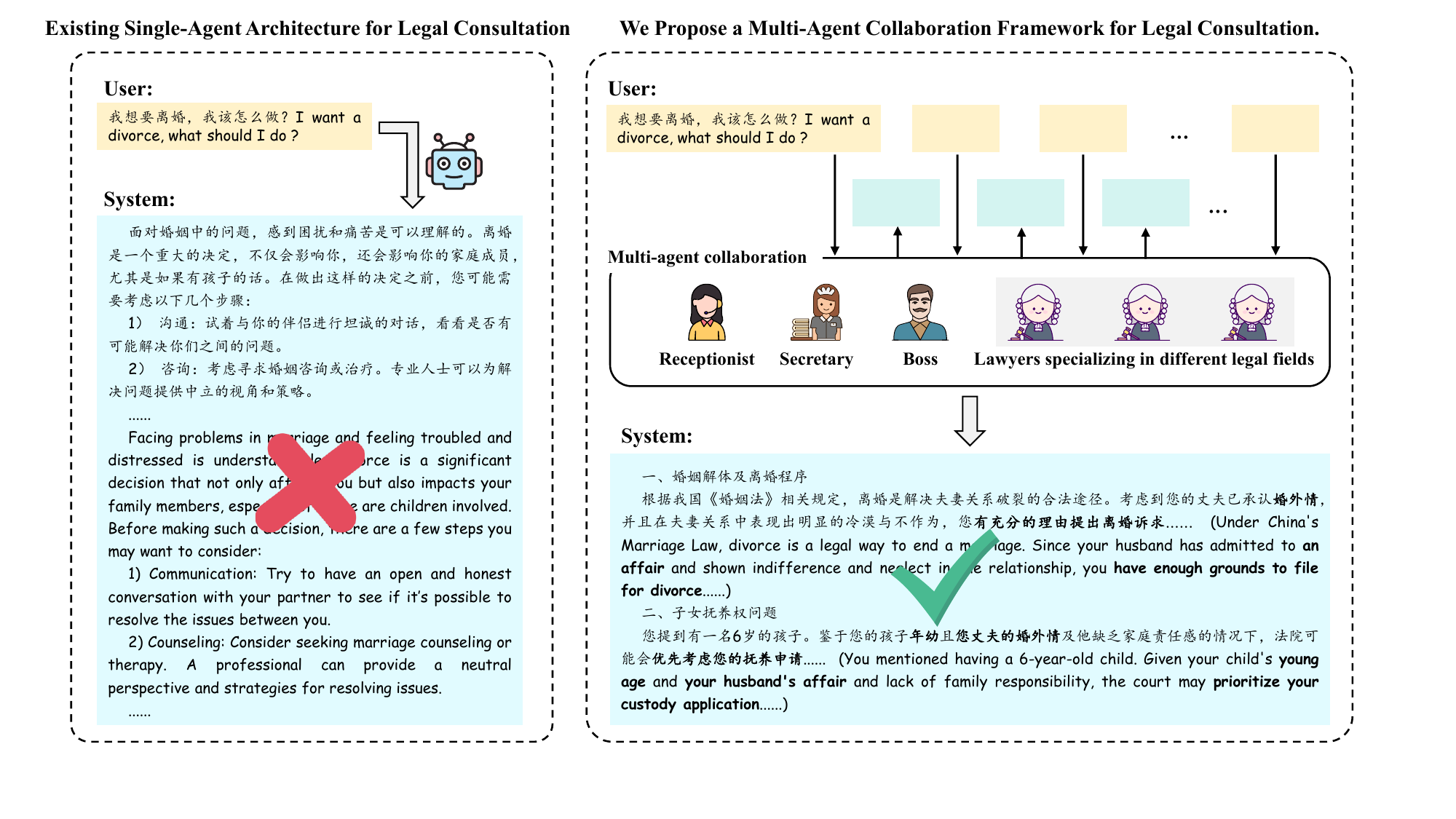}
    \caption{The left side shows the single-agent architecture used by most legal consultation systems, producing superficial, generalized responses without understanding user intent and case details. The right side presents our proposed multi-agent framework, offering more personalized and professional answers.}
    \label{Figure_1}
\end{figure*}

Recently, notable Chinese legal LLMs, such as LawGPT \citep{lawgpt}, Hanfei \citep{HanFei}, FuziMingcha \citep{sdu_fuzi_mingcha} and Lawyer-Llama \citep{lawyerllama}, have emerged. These models leverage large Chinese legal dialogue datasets to fine-tune Chinese base models, endowing them with extensive legal knowledge and the ability to engage in legal consultation dialogues. \textbf{However}, they fall short of replicating the collaborative workflows of real law firms, limiting their ability to provide personalized, professional responses, as shown in Figure \ref{Figure_1}.

To address this, we propose \textbf{LawLuo}, a multi-agent framework designed to simulate the operations of a law firm and offer professional legal advice, as shown in Figure \ref{Figure_1}. This framework consists of four distinct agents: a receptionist, a lawyer selected from the lawyer pool, a secretary, and a boss. The receptionist agent is responsible for assessing a user's intent and assigning a lawyer specializing in the relevant field. The lawyer agent analyzes the user's case and provides responses for each round of the conversation. The secretary agent organizes the entire consultation record and generates a final, personalized, and professional response for the user. The boss agent monitors the performance of both the lawyer and the secretary agents. We design a interaction strategy between these agents to simulate the operational processes of real law firms, enabling seamless collaboration to address users' legal consultations. 

To enhance the ability of each agent to follow legal instructions, we have constructed three fine-tuning datasets, including: a dataset comprising (Inquire, Lawyer description) pairs for fine-tuning the receptionist agent, a \textbf{MU}lti \textbf{R}ounds \textbf{LE}gal \textbf{D}ialogue (MURLED) dataset for fine-tuning the lawyer agent, and a \textbf{L}egal \textbf{C}onsultation \textbf{R}eport \textbf{G}eneration (LCRG) dataset for fine-tuning the secretary agent. Additionally, to address ambiguous queries, we introduce case graph-based RAG to enhance LawLuo's handling of such queries.

We evaluated LawLuo using GPT-4o and human experts. Experimental results show that LawLuo offers more personalized and professional legal advice compared to baselines. Moreover, when responding to vague questions from users without legal background, baselines often give broad answers directly. In contrast, LawLuo is committed to guiding users to clearly describe case details through leading responses. The experiments also prove LawLuo's strong ability to follow instructions even after multiple rounds of conversation.

Our primary contributions are as follows:

 \begin{itemize}
     \item We introduce a multi-agent collaborative legal dialogue framework that transcends the traditional single-model-user interaction paradigm. This innovation provides users with more personalized and professional consultation services.
     \item We constructed three different fine-tuning datasets and used them to fine-tune three different agents.
     \item We propose a case graph-based RAG to handle ambiguous queries from users without a legal background.
 \end{itemize}

\section{Related work}
\label{section:related}

\subsection{LLMs for Legal Consultation}

In recent years, large language models (LLMs) have made significant progress in various fields, particularly in the domain of Chinese law, where they have demonstrated immense potential \citep{xiao2021lawformer, yue2023disc, yang2024large}. By training on large volumes of Chinese legal case data, Legal LLMs are able to deeply understand case information and provide users with reasonable legal advice.

Most research relied on continuing pre-training and instruction fine-tuning of existing Chinese base models, aiming to enhance the models' understanding of legal knowledge and their ability to follow legal instructions \citep{lawgpt, lawyerllama, li2024csaft, dahl2024large}. Their training data mainly consists of publicly available legal documents, judicial exam data, and legal Q\&A datasets. Additionally, some studies, such as HanFei \citep{HanFei}, have opted to train a legal LLM from scratch, aiming to endow the model with more robust and profound legal knowledge and application capabilities. Some work also utilizes external legal knowledge during the reasoning phase to enhance the model's responses.\citep{louis2024interpretable, han2024rag, wan2024reformulating}

However, existing efforts have focused on improving the performance of individual legal LLMs. In practice, legal consultations in real law firms are often conducted collaboratively by multiple professionals. Inspired by this real-world work model, we propose a multi-agent collaboration framework to simulate this process, thereby providing users with a more personalized and professional legal consultation experience.

\subsection{Multi-Agent Collaboration}
In LLM-based multi-agent systems, an agent is defined as an autonomous entity capable of perceiving, thinking, learning, making decisions, and interacting with other agents \citep{xi2023rise, xu2024magic}. Research shows that breaking complex tasks into simpler subtasks and tackling these with agents that have diverse functions can significantly enhance the problem-solving capabilities of LLMs. \citep{wang2024survey, guo2024large}. For instance, \citep{qian2023communicative} designed a multi-agent collaborative workflow in which agents assuming roles such as CTO, programmer, designer, and tester work closely together to complete software development and document the development process. \citep{hemmer2022forming} have facilitated the construction of machine learning models through collaboration between multiple agents and humans. 

In addition, LLM-based multi-agent systems can also be used for simulating real-world social environments, supporting the observation and research of social behavior \citep{wang2023recagent, wei2023multi, du2023improving}.

We believe that legal consulting is a complex task that should be decomposed into subtasks, which can be collaboratively handled by multiple agents to enhance the personalization and professionalism of the responses.

\section{Framework}
\label{section:lawluo}
In real-world scenarios, legal consultations involve collaboration among multiple staff members in a law firm, while current legal LLMs engage with users in isolation. To address this gap, we propose a multi-agent collaborative framework for legal consultation, called LawLuo.

\begin{figure*}
    \centering
    \includegraphics[width=0.8\linewidth]{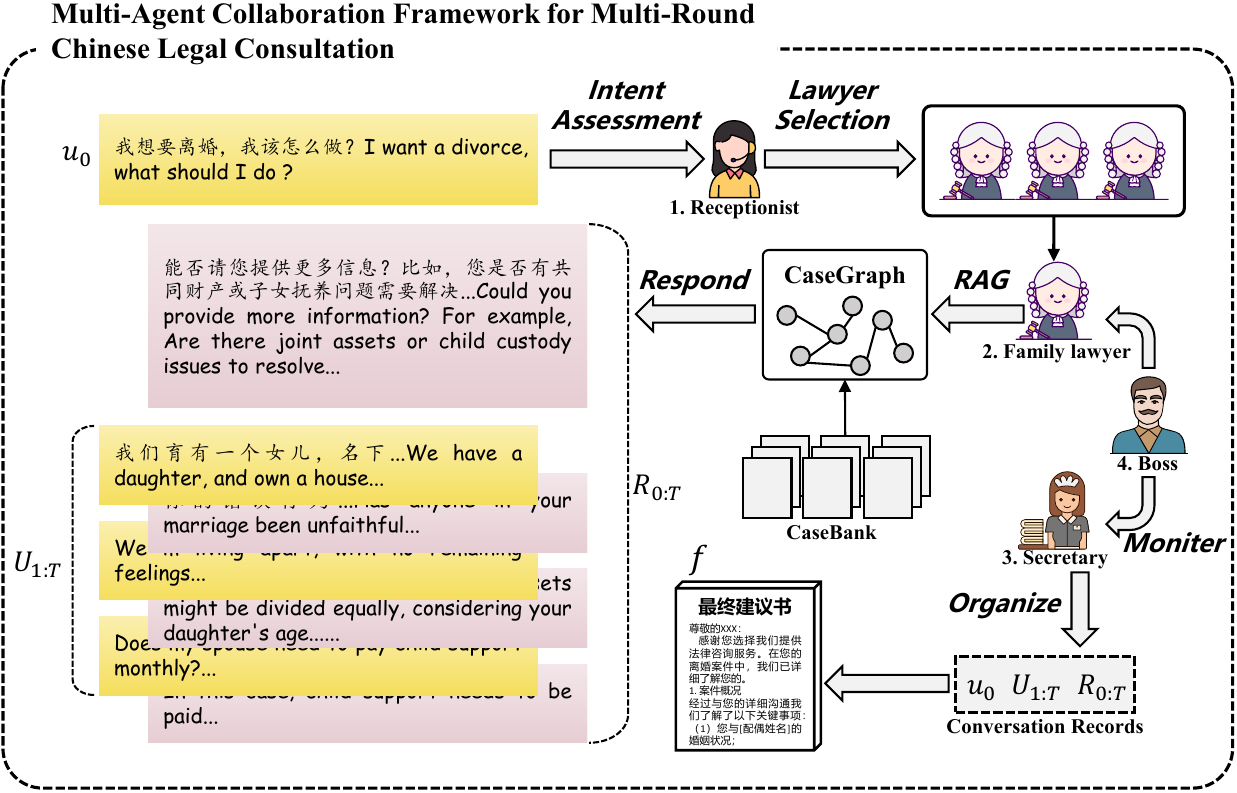}
    \caption{The multi-agent collaboration framework we propose for multi-round Chinese legal consultation. In this framework, the receptionist agent first assesses the user's consultation intent based on the initial input \( u_0 \) and selects the most suitable lawyer from the lawyer pool. Subsequently, the selected lawyer agent is responsible for engaging in multi-round dialogues with the user. During this process, the lawyer agent actively queries the user for case details via case graph-based RAG. Finally, the secretary agent organizes the dialogue records between the user and the lawyer, producing a comprehensive consultation report. The boss agent monitors the performance of the lawyer and secretary agents to ensure optimal outcomes.}
    \label{Figure_2}
\end{figure*}

The framework consists of four agent types, as shown in Figure \ref{Figure_2}: \textbf{1)} a receptionist agent, which assesses the user's consultation intent and assigns the appropriate lawyer; \textbf{2)} a lawyer agent, selected from the lawyer pool, who interacts with the user to analyze the case details; \textbf{3)} a secretary agent, which organizes the dialogue records between the lawyer and the user to generate a final consultation report; and \textbf{4)} a boss agent, which monitors the performance of both the secretary and the lawyer to ensure optimal operation.

Given the initial inquiry \( u_0 \) from the user, we will now provide a detailed description of the collaborative process of the agents within this framework.

\subsection{Receptionist}
\label{section:instruct}

Given the user's initial inquiry \( u_0 \), the receptionist agent \( R \) is tasked with evaluating the user's intent and selecting the most suitable lawyer from a pool of candidate lawyers \( \mathcal{L} \), each specializing in distinct fields, to address the user's consultation. This process is formalized in Equation \ref{eq_1}.
\begin{equation}
R: u_0 \overset{\text{max}}\mapsto \, \arg_{L \in \mathcal{L}} \, similarity(u_0, L) \label{eq_1}
\end{equation}
Where \( similarity(\cdot, \cdot) \) represents the similarity between \( u_0 \) and the description of lawyer \( L \). We defined 16 descriptions for lawyers specializing in different areas of law, based on the thematic categories of legal consultations on the HuaLv website \footnote{\href{https://www.66law.cn/}{https://www.66law.cn/}}. These areas include: Contract Law, Labor Law, Corporate Law, Intellectual Property Law, Criminal Law, Civil Procedure Law, Family Law, Real Estate Law, Tax Law, Environmental Law, Consumer Protection Law, Antitrust Law, International Trade Law, Insurance Law, Maritime Law, and others.

We constructed a dataset consisting of 1,600 pairs of (Inquire, Lawyer Description) to fine-tune the Chinese base model BaiChuan \citep{baichuan}. The fine-tuned model is employed as the receptionist agent \(R\). 

\subsection{Lawyer}
\label{section:instruct}

The lawyer agent \( L \), selected from the lawyer pool \( \mathcal{L} \), is tasked with engaging in dialogue with the user to acquire a comprehensive understanding of the case details and generate responses. This process is formally represented by Equation \ref{eq_2}.
\begin{equation}
L: (u_0, U_{1:T}) \mapsto R_{0:T} \label{eq_2}
\end{equation}
Where $U_{1:T}$ represents the sequence of user queries from the first round to the $T$-th round, $R_{0:T}$ represents the sequence of the model responses

\begin{figure}[ht]
    \centering
    \includegraphics[width=1\linewidth]{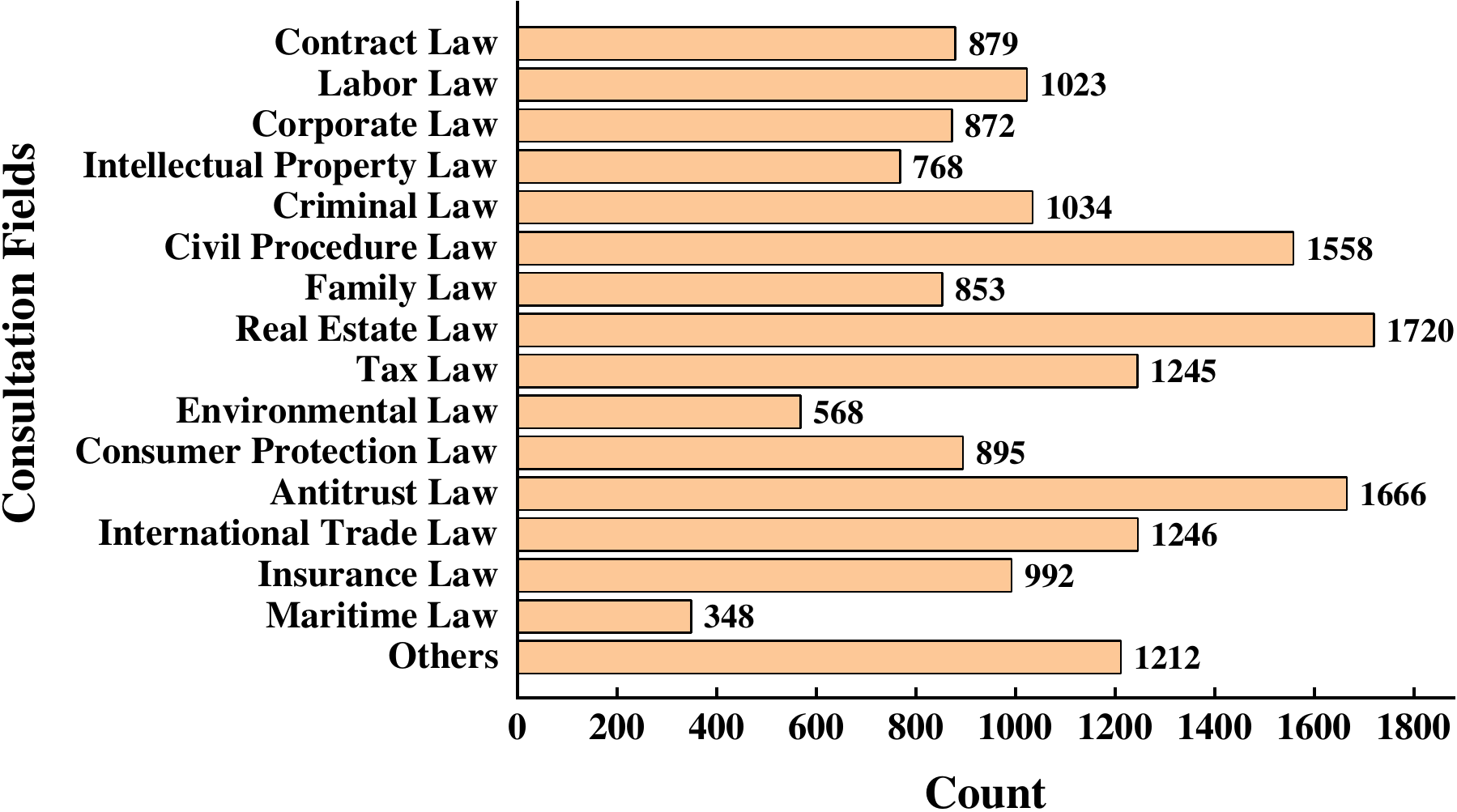}
    \caption{Distribution of the MURLED dataset across 16 different consultation domains.}
    \label{figure_3}
\end{figure}

The existing Legal LLMs, although capable of engaging in dialogue with users, tend to provide one-time responses to user queries. This contrasts with real-world legal consultations, where lawyers often engage in multiple guided conversations to gain a deeper understanding of the client's case details. To address this, we constructed a \textbf{MU}lti \textbf{R}ounds \textbf{LE}gal \textbf{D}ialogue (MURLED) dataset to fine-tune the Chinese base model ChatGLM \citep{glm}, aiming to enhance the model's legal dialogue capabilities, particularly its ability to actively guide in multiple rounds of dialogue. It is worth noting that the MURLED dataset is divided into 16 distinct consulting domains, with 16 different weight checkpoints fine-tuned on Baichuan, each serving as a lawyer agent specialized in a different consulting domain. The distribution of the MURLED dataset across 16 legal consultation fields is shown in Figure \ref{figure_3}.

\begin{figure}[ht]
    \centering
    \includegraphics[width=1\linewidth]{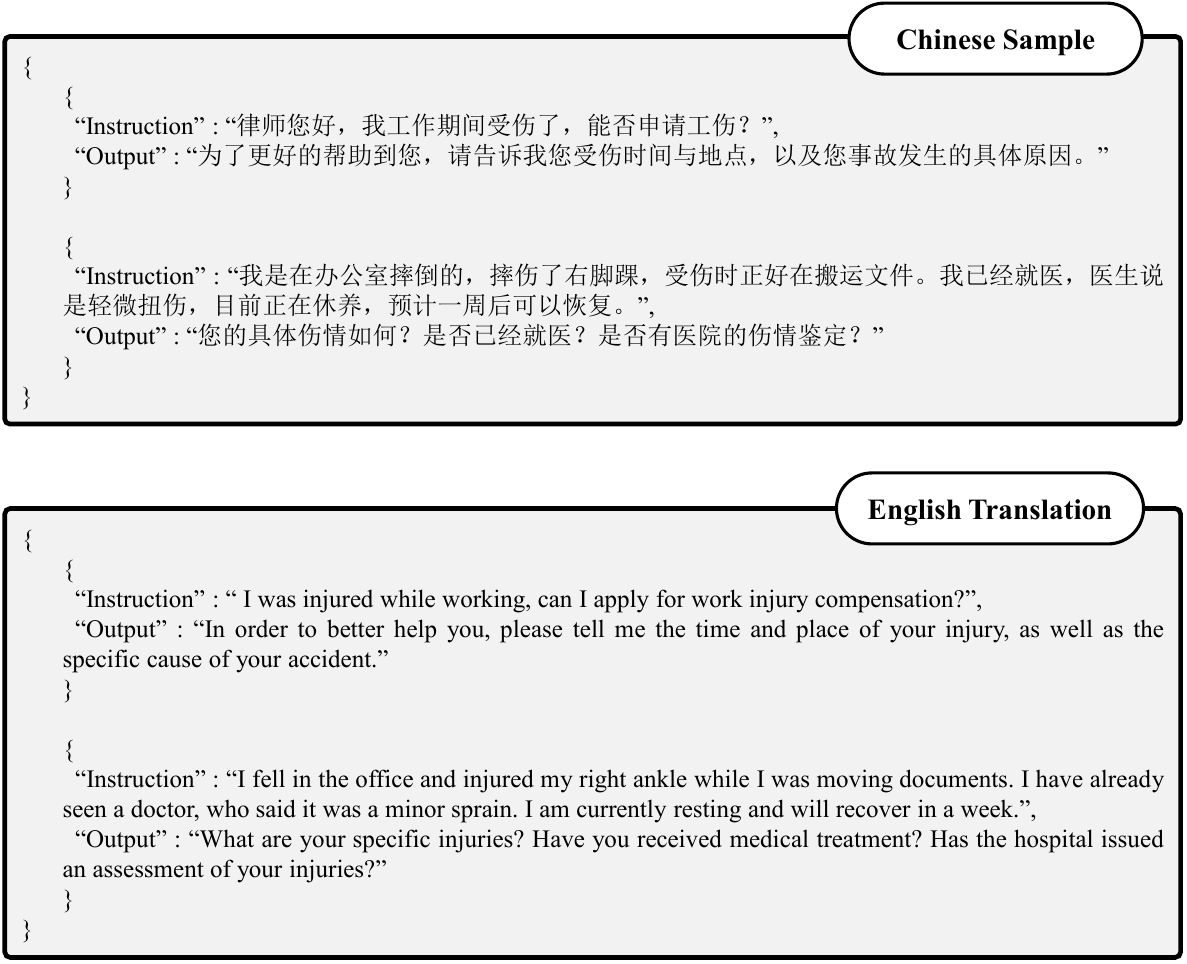}
    \caption{An example from the MURLED dataset. It can be seen that this dataset emphasizes the active guidance ability of training large legal models in multi-turn dialogues.}
    \label{figure_4}
\end{figure}

The MURLED dataset was constructed based on case consultation voice recordings from a law firm and contains 16,734 multi-turn legal conversations. We first converted the raw audio files into text format and then utilized ChatGPT to optimize the text, addressing issues such as informality and minor errors. Figure \ref{figure_4} shows an example of a multi-turn conversation from MURLED, highlighting the emphasis on guiding responses.

We used MURLED to fine-tune ChatGLM-3-6b. To mitigate the risk of overfitting, we incorporated general conversational data from Alpaca-GPT4\footnote{\href{https://www.modelscope.cn/datasets/AI-ModelScope/alpaca-gpt4-data-zh/summary}{https://www.modelscope.cn/datasets/AI-ModelScope/alpaca-gpt4-data-zh/summary}}, which comprises 52,000 generic Chinese dialogues, into the fine-tuning process. To expedite the fine-tuning of the model and reduce reliance on computational resources, we employed the LoRA fine-tuning strategy, as illustrated in Equation \ref{eq_3}:
\begin{equation}
\theta_{Legal} = LoRA \left( \theta, \left\{ \left( u_0^n, U_{1:T}^n, R_{0:T}^n \right)_{n=1}^N \right\} \right) \label{eq_3}
\end{equation}
where \(\theta\) represents the initial parameters of ChatGLM-3-6b, while \(\theta_{Legal}\) denotes the parameters of our fine-tuned legal LLM. Besides, \((u_0^n, U_{1:T}^n, R_{0:T}^n)\) indicates the \(n\)-th training sample.

To enhance the lawyer agent's ability to address vague queries from users without a legal background, we design the agent to employ a case graph-based Retrieval-Augmented Generation (RAG) approach during each response generation process. Specifically, we implement this case graph-based RAG using the LightRAG framework \citep{guo2024lightrag}. To build the case graph, we utilize a case collection comprising 4,320 criminal cases and 12,345 civil cases, sourced from the China Judgments Online database \footnote{\href{https://wenshu.court.gov.cn/}{https://wenshu.court.gov.cn/}}. The construction of the case graph is detailed in Algorithm \ref{alg:1}.

\begin{algorithm}
\caption{Case Graph Construction}
\begin{algorithmic}[1]
\State \textbf{Input:} Set of legal cases $\mathcal{C} = \{c_1, c_2, \dots, c_n\}$
\State \textbf{Output:} Case graph $G = (V, E)$

\For{each case $c_i$ in $\mathcal{C}$}
    \State $v_i \gets f(c_i)$  \Comment{Generate vector representation of case $c_i$}
\EndFor
\State \textbf{For each pair of cases:}
\For{each $c_i, c_j \in \mathcal{C}$}
    \State $Sim(c_i, c_j) \gets \text{similarity}(v_i, v_j)$  \Comment{Compute similarity between cases}
    \State Add edge $(c_i, c_j, Sim(c_i, c_j))$ to $E$  \Comment{Add weighted edge to graph}
\EndFor
\State \textbf{Return:} Case graph $G = (V, E)$
\end{algorithmic}
\label{alg:1}
\end{algorithm}

\subsection{Secretary}
\label{section:instruct}

The secretary agent's responsibility is to organize the conversation records between the user and the lawyer, and compile a final consultation report to be submitted to the user, as shown in Equation \ref{eq_4}.

\begin{equation}
S:(u_0,U_{1:T},R_{0:T} ) \mapsto f
\label{eq_4}
\end{equation}
Where \( f \) represents the final consulting report.

We created a \textbf{L}egal \textbf{C}onsultation \textbf{R}eport \textbf{G}eneration dataset called \textbf{LCRG}, which includes 420 legal consultation dialogues and their summary reports. Each summary report is carefully written by professional lawyers. We used LCRG to fine-tune the Chinese base model BaiChuan, enabling the model to generate consultation reports from legal consultation dialogues. A legal consultation summary report sample is shown in Appendix \ref{appendix_a}.

\subsection{Boss}
\label{section:instruct}

The boss agent is responsible for evaluating and optimizing the performance of the lawyer and secretary agents. We treat the boss agent as a binary reward model, $B: o \mapsto y$, where $o$ represents the output of the lawyer or secretary agent, and $y$ represents the evaluation of $o$ by the boss agent, categorized as "better" or "worse." The training objective for the boss agent is to minimize the following loss function:
\begin{equation}
\begin{aligned}
\mathcal{L}_{B} = -\frac{1}{N} \sum_{i=1}^{N} & \left[ y_i \cdot \log \left( \hat{y}_i(o_i; \theta_{B}) \right) \right. \\
& \left. + (1 - y_i) \cdot \log \left( 1 - \hat{y}_i(o_i; \theta_{B}) \right) \right]
\end{aligned}
\label{eq:loss_rm}
\end{equation}
where \( y_i \) represents the true label of the \( i \)-th sample, taking values of either 0 or 1, which correspond to ``worse'' and ``better'', respectively. Besides, \( \hat{y}_i (o_i; \theta_{B}) \) denotes the probability that boss predicts the \( i \)-th output \( o_i \) as ``better''.

We adopt the PPO algorithm \citep{wang2020truly} to enable reinforcement learning between the boss agent and the lawyer agent, as well as between the boss agent and the secretary agent. Through this reinforcement learning, the boss agent continuously optimizes the lawyer and secretary agents.

\section{Experimental Setup}
\label{section:exper}

All our experiments were conducted on a 40G A100 GPU. The PyTorch 2.3.0 and the HuggingFace Transformers 4.40.0 were used. The learning rate for LoRA fine-tuning was set to 0.00005, with a training batch size of 2, over a total of 3 epochs, and model weights were saved every 1,000 steps. Additionally, the rank of LoRA was set to 16, the alpha parameter was set to 32, and the dropout rate was set to 0.05.

\begin{figure*}
    \centering
    \includegraphics[width=1\linewidth]{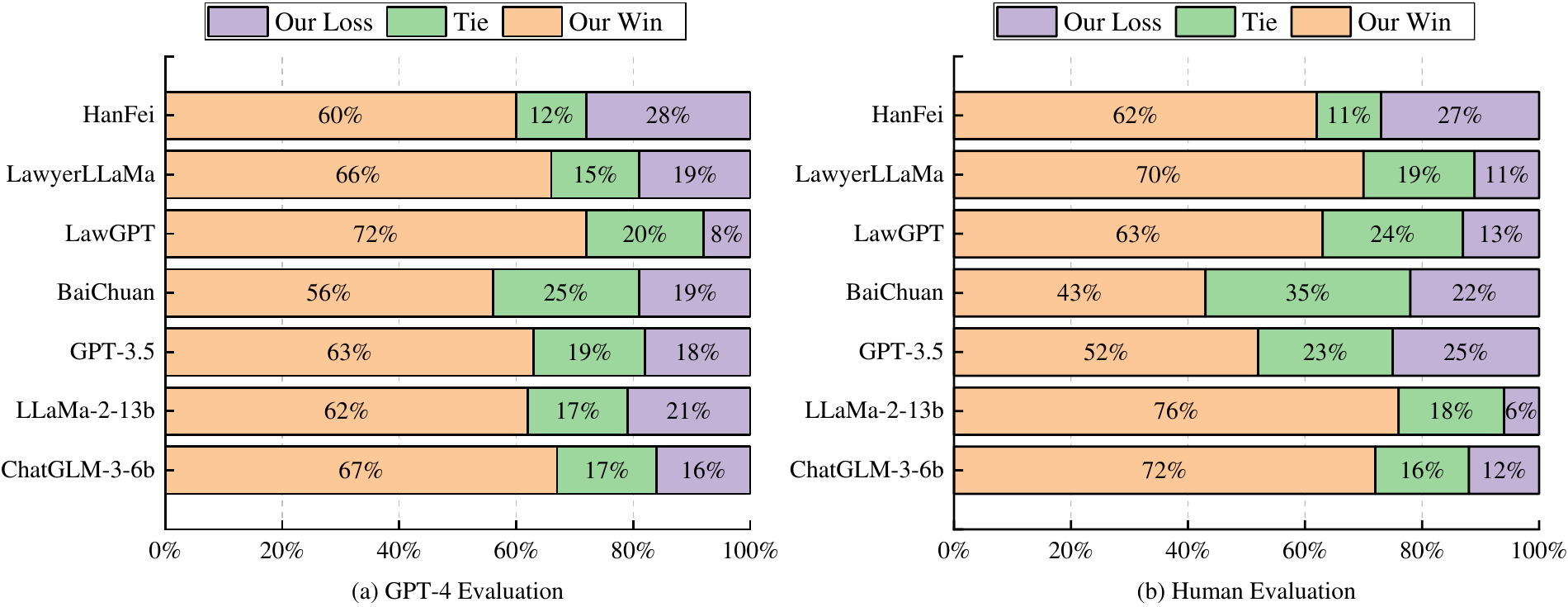}
    \caption{Win rate of LawLuo compared to the baselines}
    \label{mainresults}
\end{figure*}

\section{Results and Analysis}
\label{section:result}

The following research questions will be addressed through experimental analysis:

\textbf{RQ 1}: Does multi-agent collaboration facilitate the generation of more personalized and professional responses for users' legal consultations?

\textbf{RQ 2}: Can LawLuo more effectively address legal inquiries raised by users without a legal background?

\textbf{RQ 3}: After engaging in multiple rounds of legal dialogue, can LawLuo maintain its ability to follow legal instructions accurately?

\textbf{RQ 4}: Does the instruction fine-tuning applied to the constructed datasets improve the performance of the agents?

\textbf{RQ 5}: Is LawLuo still effective in performing routine legal tasks, including non-dialogue tasks?

\subsection{Pairwise Comparison Evaluation}
\label{section:single}

We employed pairwise comparison to assess the performance of LawLuo. In the evaluation, the outputs generated by LawLuo is compared with the outputs generated by the baselines  using the same input data, in terms of personalization and professionalism, by GPT-4 or human experts. For each comparison, experts are asked to determine whether LawLuo performs better, worse, or similarly to the baseline models. This evaluation method is consistent with current best practices for evaluating large language models \citep{thirunavukarasu2023large, xiong2023doctorglm, zhang2023huatuogpt}.

Figure \ref{mainresults} presents the win rate of LawLuo against the baselines, clearly showing that LawLuo outperforms widely used Chinese base LLMs and exceeds all legal LLM baselines. This results answer \textbf{RQ 1}: Collaboration among multiple agents in legal consultation can indeed provide users with more personalized and professional responses.

\subsection{Case Study on Ambiguous Inquiry}
\label{section:multi}

We randomly select a ambiguous legal consultation question and analyze the answers generated by LawLuo, ChatGLM-3, BaiChuan, LawGPT, and HanFei in the first round, as shown in Table \ref{tab:2}. From the table, it can be seen that LawLuo's responses in the first round are more guiding. This guiding response helps users to better elaborate on the case details, thereby providing the most personalized and accurate answers. The experimental results address \textbf{RQ 2}: LawLuo is better at handling ambiguous legal consultations from users without a legal background.

\subsection{Multi-Turn Dialogue}
\label{section:case}

\begin{figure}
    \centering
    \includegraphics[width=1\linewidth]{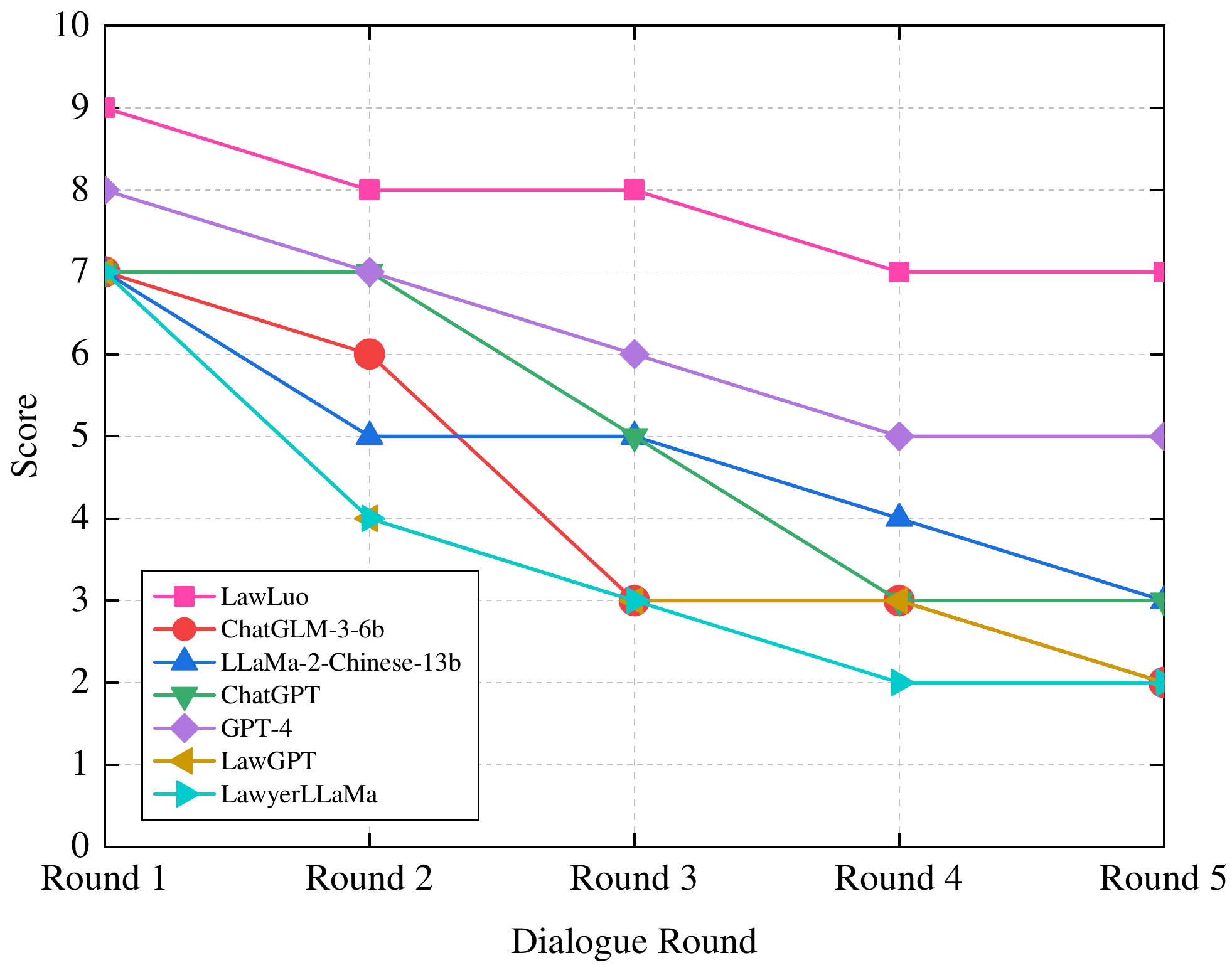}
    \caption{The variation in the quality of model-generated responses with increasing dialogue turns}
    \label{multiresults}
\end{figure}

We systematically evaluate the instruction-following capability of the proposed LawLuo model in multi-turn dialogues. The experimental design includes four dialogue scenarios where instructions evolve or become progressively more complex across turns. We assess the model’s ability to understand and execute instructions through tasks such as legal charge prediction, similar case matching, and case element extraction. The evaluation metrics focus on the model's accuracy in understanding instructions, coherence in maintaining context, precision in execution, adaptability, flexibility, and its ability to handle complex or conflicting instructions. We use GPT-4 to evaluate LawLuo and the baselines' instruction-following scores after multiple rounds of dialogue, as detailed in Appendix \ref{appendix_c}. The experimental results are shown in Figure \ref{multiresults}, with the pink line representing LawLuo's instruction compliance score. It can be observed that even after five rounds of dialogue, LawLuo still maintains a high level of instruction compliance. This experimental result answers \textbf{RQ3}: LawLuo is still able to effectively comply with legal instructions after multiple rounds of dialogue.

\subsection{Ablation Study}
\label{section:ablation}

\begin{figure}
    \centering
    \includegraphics[width=1\linewidth]{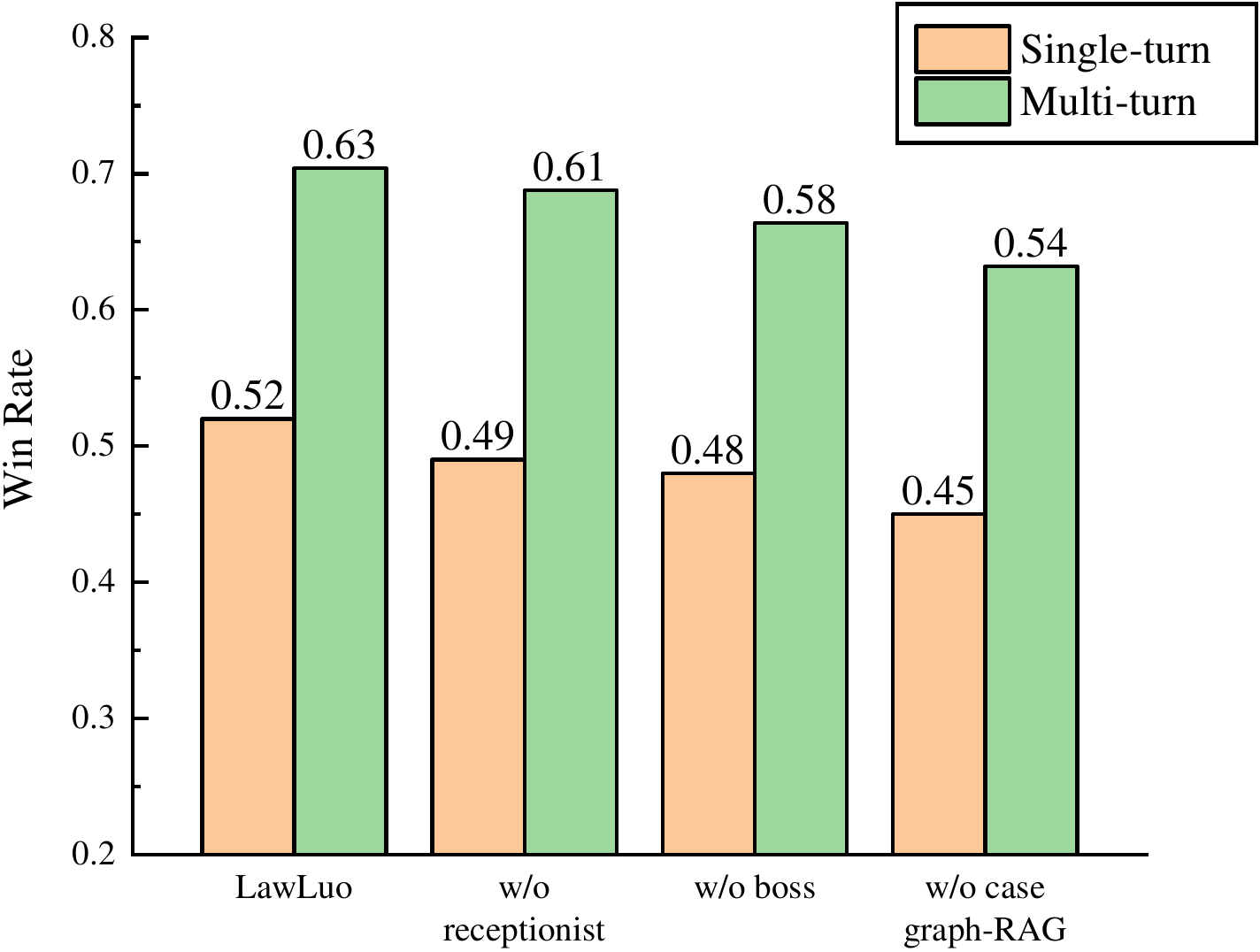}
    \caption{Results of ablation experiments}
    \label{ablation}
\end{figure}

This section aims to validate the contributions of each component within the framework. We continue to use GPT-4o as the evaluator to assess the win rate of LawLuo over GPT-3.5 after ablation, as illustrated in Figure \ref{ablation}. From the figure, it is evident that the win rate of LawLuo over GPT-3.5 decreases by 2\% after ablating the receptionist agent. This result validates our hypothesis that legal LLMs should be assigned different domain-specific roles to provide more targeted answers based on the user's consultation field. Additionally, the figure shows that the boss agent also contributes to LawLuo's performance, as it can optimize the responses generated by the lawyer. Finally, we observe a significant decline in model performance after removing the case graph-based RAG module. This indicates that clarifying users' vague and ambiguous queries is crucial for generating high-quality responses in legal question-answering. The experimental results answer \textbf{RQ4}: Our fine-tuning of each agent enables LawLuo to achieve better overall performance.

\begin{table*}[ht]
\centering

\caption{Performance of LawLuo and the baselines on five routine legal natural language processing tasks, reflecting their understanding of legal knowledge.}
\resizebox{\linewidth}{!}{
\begin{tabular}{@{\hskip 5pt}c@{\hskip 5pt}|@{\hskip 5pt}c@{\hskip 5pt}|@{\hskip 5pt}c@{\hskip 5pt}|@{\hskip 5pt}c@{\hskip 5pt}|@{\hskip 5pt}c@{\hskip 5pt}|@{\hskip 5pt}c@{\hskip 5pt}}
\toprule
\textbf{Task} & \shortstack{Legal Event \\ Extraction \\ (F1 score)} & \shortstack{Judicial Reading \\ Comprehension \\ (F1 score)} & \shortstack{Legal Charges \\ Prediction \\ (F1 score)} & \shortstack{Related Law \\ Retrieval \\ (F1 score)} & \shortstack{Similar Case \\ Retrieval \\ (Acc@10)} \\
\midrule
\textbf{Dataset} & LEVEN \citep{yao2022leven} & CJRC \citep{duan2019cjrc} & CAIL2018 \citep{xiao2018cail2018} & CAIL2018 \citep{xiao2018cail2018} & LeCaRD \citep{ma2021lecard} \\
\midrule
LawLuo         & 73.3 $\pm$ 1.3 & 80.6 $\pm$ 2.3 & 92.1 $\pm$ 2.3 & 84.4 $\pm$ 3.1 & 81.6 $\pm$ 3.8 \\
HanFei         & \textbf{73.5 $\pm$ 1.2} & \textbf{83.2 $\pm$ 1.7} & 92.1 $\pm$ 2.2 & \textbf{84.5 $\pm$ 2.0} & 83.0 $\pm$ 3.1 \\
LawGPT         & 72.5 $\pm$ 1.3 & 81.5 $\pm$ 2.1 & 90.8 $\pm$ 1.5 & 83.2 $\pm$ 2.8 & \textbf{85.5 $\pm$ 3.0} \\
LawyerLLaMa    & 71.2 $\pm$ 1.4 & 80.2 $\pm$ 2.2 & 91.7 $\pm$ 2.3 & 81.6 $\pm$ 3.1 & 82.1 $\pm$ 3.5 \\
BaiChuan       & 72.0 $\pm$ 1.3 & 82.2 $\pm$ 1.5 & 92.5 $\pm$ 2.3 & 83.6 $\pm$ 3.4 & \textbf{85.5 $\pm$ 2.9} \\
ChatGLM-3-6b   & 72.2 $\pm$ 1.4 & 79.8 $\pm$ 2.2 & \textbf{94.4 $\pm$ 2.3} & 82.1 $\pm$ 2.8 & 84.4 $\pm$ 3.3 \\
GPT-3.5        & 70.9 $\pm$ 2.1 & 77.6 $\pm$ 2.4 & 90.5 $\pm$ 2.4 & 81.1 $\pm$ 2.2 & 80.4 $\pm$ 3.2 \\
\bottomrule
\end{tabular}
}
\label{tab:1}
\end{table*}

\subsection{Legal Knowledge Probing Experiment}

We evaluated LawLuo's performance across five routine legal natural language processing tasks: Legal Event Extraction, Judicial Reading Comprehension, Legal Charges Prediction, Related Law Retrieval, and Similar Case Retrieval. These tasks were conducted on established datasets, including LEVEN \citep{yao2022leven}, CJRC \citep{duan2019cjrc}, CAIL2018 \citep{xiao2018cail2018}, and LeCaRD \citep{ma2021lecard}.

The results, as presented in Table \ref{tab:1}, indicate that LawLuo performs well across all five tasks. Although its performance is not the best when compared to other baseline models, it remains highly competitive. This suggests that through instruction fine-tuning, LawLuo has acquired sufficient legal knowledge, enabling it to not only handle legal consultations but also address routine legal natural language processing tasks. The experimental results answer \textbf{RQ5}: LawLuo remains effective in routine legal tasks.

\section{System Implementation}

\begin{figure}
    \centering
    \includegraphics[width=1\linewidth]{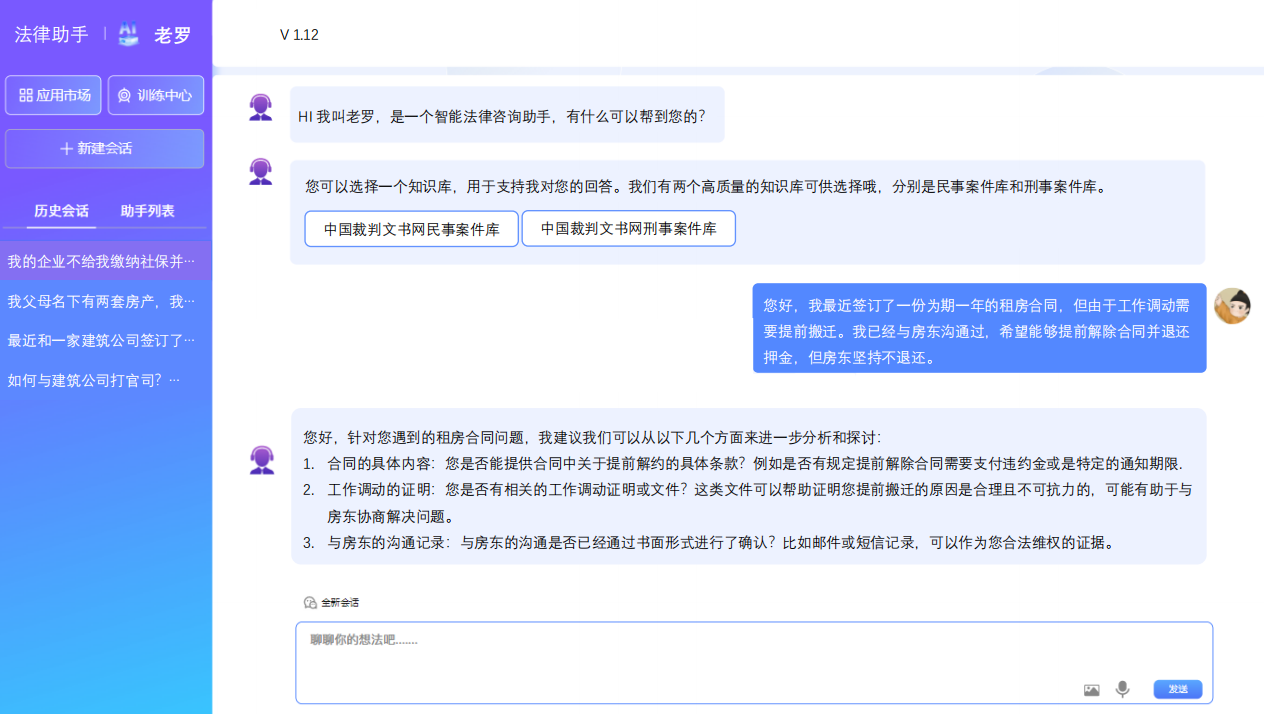}
    \caption{We have built a web-based legal consultation system with the LawLuo framework as the core, and testing has shown that it has good practical effectiveness.}
    \label{sys}
\end{figure}

Based on the LawLuo framework, we have designed and implemented a practical legal consultation system aimed at providing users with an efficient and interactive legal advisory platform, as shown in Figure \ref{sys}. The system's backend is developed using the Flask framework, while the frontend is built with React to ensure a dynamic and responsive user experience. Users access the system through a simple web interface and initially interact with a receptionist agent to describe their legal issues. The system then guides users to the relevant lawyer agent for a detailed case discussion. After several rounds of conversation, the secretary agent generates and provides a legal consultation report, while the boss agent monitors the entire interaction process in the background to ensure service quality.

\section{Conclusion}

We introduce LawLuo, a multi-agent collaboration framework that simulates the multi-party interactions of real law firms to provide professional legal consulting services. Experimental results demonstrate that LawLuo outperforms traditional single-agent models in generating personalized and professional legal advice, handling ambiguous inquiries, and following legal instructions in multi-turn dialogues. Ablation studies and legal knowledge probing experiments further validate the effectiveness of various components within the framework, as well as the legal knowledge acquired through instruction tuning. Despite these achievements, we acknowledge that there is room for improvement in optimizing inter-agent communication and enhancing model interpretability, which will be the focus of future research. The successful implementation of LawLuo paves the way for new developments in the field of legal consulting, suggesting the broad application prospects of multi-agent collaboration in future legal services.

\section*{Limitation and Future Work}

The experimental outcomes of the LawLuo framework underscore the potential of multi-agent collaboration within the domain of legal consultation. By emulating the multi-party interactions characteristic of real law firms, our model is capable of delivering consultation service that are more personalized and professional. The strength of this collaborative approach lies in its ability to comprehensively understand user needs from various perspectives and provide solutions on multiple levels. However, multi-agent systems also introduce new challenges, particularly in terms of communication and coordination among agents. To ensure seamless collaboration, each agent must possess a high degree of domain-specific expertise and be able to comprehend the decisions and feedback of other agents. Future work should further explore how to optimize the interaction mechanisms between agents to reduce misunderstandings and enhance collaboration efficiency.

\bibliography{custom}

\appendix

\section{A sample from Legal Consultation Report Generation}
\label{appendix_a}

\begin{figure*}[!h]
    \centering
    \includegraphics[width=1\linewidth]{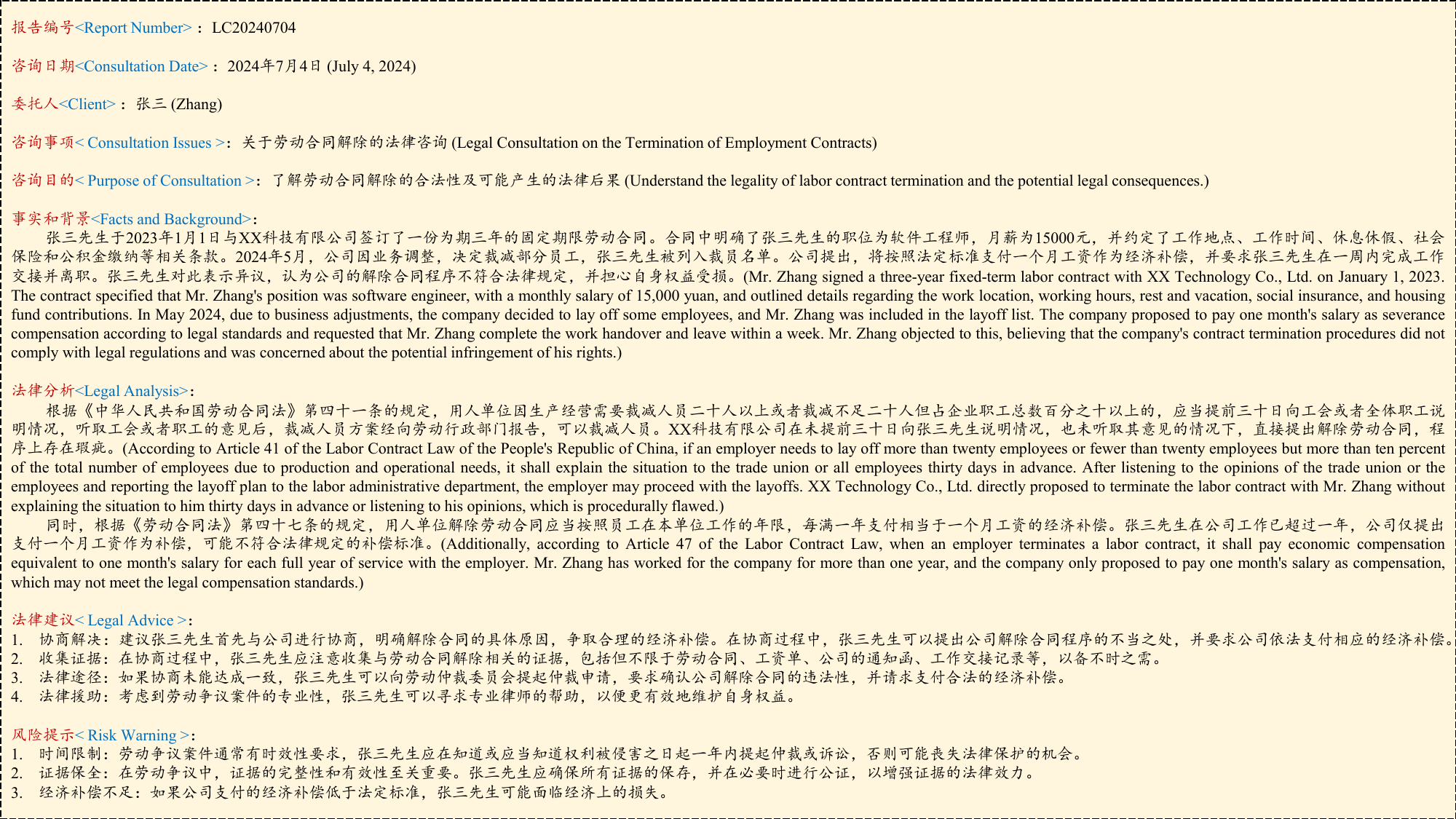}
    \caption{A sample consulting report for in-context learning}
    \label{samplereport}
\end{figure*}

Figure \ref{samplereport} is a sample of legal consultation summary report. A summary report comprises nine sections: report number, consultation date, client, subject of consultation, purpose of consultation, facts and background, legal analysis, legal advice, and risk warnings.

\onecolumn

\section{Case Analysis Presentation}
\label{appendix_b}

We present the responses of LawLuo and baselines to users' vague legal queries in Table \ref{tab:2}.

\begin{table*}[h]
\centering
\begin{tabular}{|>{\centering\arraybackslash}m{2cm}|m{12cm}|}
\hline
          & Inquiry: \begin{CJK}{UTF8}{gbsn}我的朋友借了我一些钱，现在他说他没有能力还款，我该怎么办？\end{CJK}(My friend borrowed some money from me, and now they say they are unable to repay. What should I do?)\\ \hline
LawLuo    & Response: \begin{CJK}{UTF8}{gbsn}为了更好地帮助您解决问题，能否提供更多细节？比如借款金额、还款期限，是否有书面协议或相关证据（如转账记录），朋友的财务状况，以及您与朋友之间的沟通情况（是否讨论过分期还款等方案）。这些信息有助于我们更准确地分析您的问题并提出合适的建议。\end{CJK}(To better assist you in resolving the issue, could you provide more details? For example, the loan amount, repayment term, whether there is a written agreement or related evidence (such as transfer records), your friend's financial situation, and the communication between you and your friend (whether installment repayment options were discussed, etc.). This information will help us analyze your situation more accurately and provide appropriate advice.)\\ \hline
ChatGLM-3 & Response: \begin{CJK}{UTF8}{gbsn}您可以和朋友协商解决，比如分期还款或延期。若协商不成，您可以通过法律途径，提起诉讼。\end{CJK}(You can negotiate with your friend to find a solution, such as agreeing on installment payments or a delay in repayment. If negotiation fails, you can pursue legal action and file a lawsuit.)\\ \hline
BaiChuan  & Response: \begin{CJK}{UTF8}{gbsn}遇到朋友无法还款的情况，可以先与对方沟通，了解其实际经济状况，商定双方都可以接受的还款方式。如果协商未果，建议您咨询当地法律顾问，了解您所在地区关于债务纠纷的具体法律规定。\end{CJK}(If you encounter a situation where your friend is unable to repay the loan, you can first communicate with them to understand their actual financial situation and agree on a repayment method that both parties can accept. If negotiation fails, it is advisable to consult a local legal advisor to understand the specific legal regulations regarding debt disputes in your area.)\\ \hline
LawGPT    & Response: \begin{CJK}{UTF8}{gbsn}建议首先与朋友保持良好沟通，尽量避免矛盾升级。您可以考虑灵活处理，比如延期或分期还款。如果问题无法解决，可以查阅相关法律条文或寻求律师帮助，以确保自己的合法权益得到保障。\end{CJK}(It is advisable to maintain good communication with your friend first, in order to avoid escalating conflicts. You can consider flexible solutions, such as delaying or agreeing on installment payments. If the issue cannot be resolved, you can consult relevant legal provisions or seek help from a lawyer to ensure your legal rights are protected.)\\ \hline
HanFei    & Response: \begin{CJK}{UTF8}{gbsn}此类问题的处理需要结合具体情况分析。建议您先尝试与朋友协商，充分了解其还款困难的原因，或制定双方认可的分期还款计划。如果协商无果，可以准备相关证据（如借条、转账记录等），通过正式的法律途径解决。同时，在追求法律手段前，需评估其执行能力和成本效益。\end{CJK}(The resolution of such issues requires analyzing the specific situation. It is recommended to first try negotiating with your friend, understanding the reasons behind their repayment difficulties, or creating an installment plan both parties agree on. If negotiations fail, you can gather relevant evidence (such as loan receipts or transfer records) and pursue legal action. However, before opting for legal means, it is important to evaluate the enforceability and cost-effectiveness of this approach.)\\ \hline
\end{tabular}
\caption{Responses to Inquiry about Loan Repayment Issue}
\label{tab:2}
\end{table*}

\section{Prompt Template for Guiding GPT-4o in Scoring}
\label{appendix_c}
\begin{figure*}[!h]
    \centering
    \includegraphics[width=0.8\linewidth]{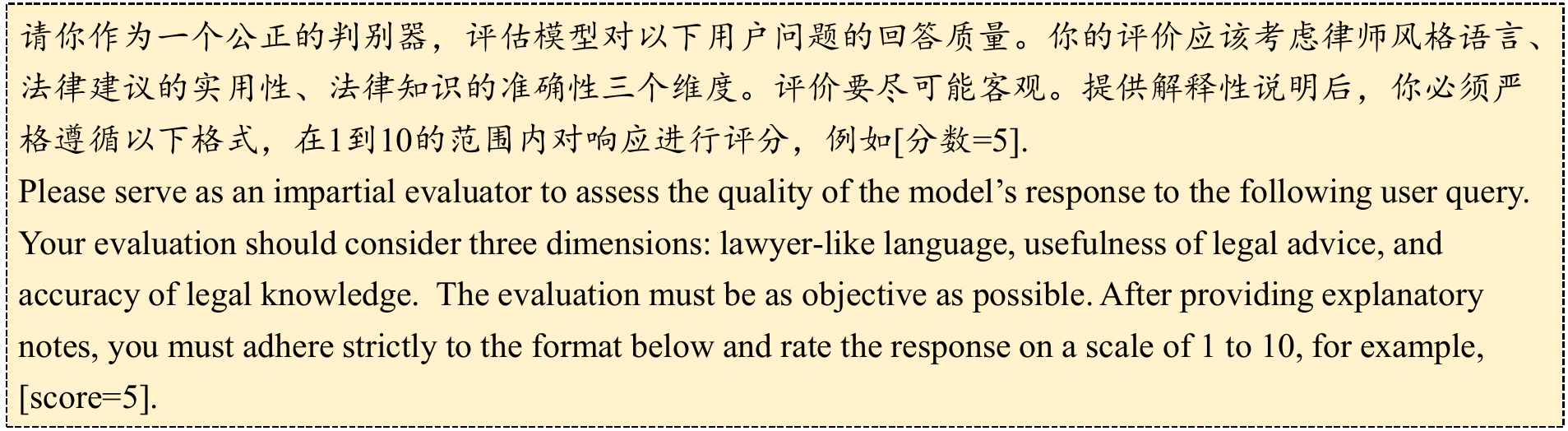}
    \caption{Prompt template for guiding GPT-4o in scoring}
    \label{scorer}
\end{figure*}

Figure \ref{scorer} is a prompt template for guiding GPT-4o in scoring.

\end{document}